# BSG4Bot:Efficient Bot Detection based on Biased Heterogeneous Subgraphs

Hao Miao[†], Zida Liu[†], and Jun Gao[†*]
[†]Key Laboratory of High Confidence Software Technologies, CS, Peking University, China
miaohao@stu.pku.edu.cn, zida-liu@hotmail.com, gaojun@pku.edu.cn

*Abstract*—The detection of malicious social bots has become a crucial task, as bots can be easily deployed and manipulated to spread disinformation, promote conspiracy messages, and more. Most existing approaches utilize graph neural networks (GNNs) to capture both user profile and structural features, achieving promising progress. However, they still face limitations including the expensive training on large underlying graph, the performance degradation when "similar neighborhood patterns" assumption preferred by GNNs is not satisfied, and the dynamic features of bots in a highly adversarial context.

Motivated by these limitations, this paper proposes a method named BSG4Bot with an intuition that GNNs training on <u>B</u>iased <u>S</u>ub<u>G</u>raphs can improve both performance and time/space efficiency in bot detection. Specifically, BSG4Bot first pre-trains a classifier on node features efficiently to define the node similarities, and constructs biased subgraphs by combining the similarities computed by the pre-trained classifier and the node importances computed by Personalized PageRank (PPR scores). BSG4Bot then introduces a heterogeneous GNN over the constructed subgraphs to detect bots effectively and efficiently. The relatively stable features, including the content category and temporal activity features, are explored and incorporated into BSG4Bot after preliminary verification on sample data. The extensive experimental studies show that BSG4Bot outperforms the state-of-the-art bot detection methods, while only needing nearly 1/5 training time.

*Index Terms*—Graph Neural Networks, Social Bot Detection, Biased subgraphs

## I. INTRODUCTION

Social bot detection, as a critical kind of outlier detection in social networks [1], has drawn increasing attention due to serious harm to user interests. The bots can be programmed and deployed in networks in a cost-efficient way to undertake specific tasks with malicious purposes in general, like spreading misinformation [2], manipulating public sentiment [3], and even interfering in political processes [4]. The proliferation of these bots poses severe damage to the integrity of information flow on social media.

The existing methods can be categorized roughly according to different models used. The early studies mainly employ traditional classifiers like Random Forests to distinguish bots from genuine users based on various features, including user metadata [5], tweet content [6], and interaction patterns [7]. Bot manipulators began to craft these features meticulously in order to bypass these detection methods. Subsequently, researchers employ deep learning models, such as Transformers and BERT, to extract implicit features from textual content [8] and descriptions [9], aiming to counteract increasingly sophisticated bots. Recently, the focus of research has shifted towards graph-based methods [10], [11], [12], [13], [14]. These methods model the network as a graph, usually utilize GNNs to capture user features and the structure of the graph, and achieve performance improvements in bot detection. In fact, the social bot detection shares similarities with other GNN-based tasks like recommendation [15], knowledge inference [16], etc. The advances of one task can benefit other similar tasks.

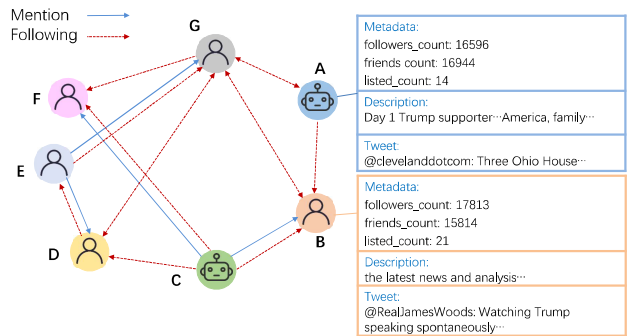

Fig. 1. An example of bot detection. Node $A$ and $C$ represent bots, and the others are genuine users. Nodes with different labels can share similar user features, and two kinds of users have slightly different structural patterns, in which genuine users are typically interconnected, whereas bots exhibit few connections among themselves but extensively link to genuine users.

Despite successes in the graph-based bot detection methods, they still face the following limitations. First, the prevalent bot detection methods [10], [11], [12], [13], [14] attempt to achieve high performance by training over the entire graph, while the underlying graphs are usually large. For instance, the Twibot-22 benchmark [17] comprises 1,000,000 nodes and 3,743,634 edges. Training a model over such a graph requires substantial computational resources, as sophisticated graph learning requires loading data into scarce GPU memory. One approach is to leverage the strategy of subgraph training [18], [19] to lower the GPU memory consumption.

Second, the existing studies show that the GNN models have the potential to achieve good performance if nodes with the same label share "similar neighborhood patterns" [20], [21]. In other words, classical GNN models may work well to handle fully homophilic or heterophilic graph solely, but face performance degradation when handling the mixture cases for different nodes [20], [21]. Taking Figure 1 as an example. The mixed structural patterns for bot and genuine users impact the performance of GNN models.

Third, bot manipulators can be aware of the detection rules, and strive to craft bot metadata and mimic the tweet content in the contest between bot design and detection. As shown in Figure 1, bot $A$, through well-designed features, mimics the characteristics of a genuine user $B$, thereby confusing the detector. We believe that without input from the experts, it is hard for GNNs with learned implicit patterns to beat human-designed bot policies. Delving deeper into the features that facilitate the distinction between bots and genuine users can also provide some hints for the model interpretability.

In this paper, we propose BSG4Bot, a framework that builds biased homophilic subgraphs with multi-relations for bot detection, to overcome above limitations. The contributions of our method are summarized as follows:

- BSG4Bot follows the subgraph training strategy to handle the large graph to lower the memory demands significantly. In addition, BSG4Bot considers heterogeneous relationships, and combines the hidden representations in different layers with semantic attention for better performance.
- BSG4Bot proposes a biased subgraph construction method which is likely to select neighboring nodes with the similar labels to the starting node of the subgraph. Specifically, BSG4Bot pre-trains a coarse classifier on node features only, and builds subgraphs by combining the similarities to the start node and importances computed by PPR. Such a strategy can enhance the subgraph homophily, which is favoured by GNN models to boost the performance.
- We conduct a preliminary data observation over the existing data, and extract potential distinguishable features from the viewpoints of content and temporal behaviors of users. These features are incorporated into BSG4Bot and further verified in the ablation experimental study.

We perform extensive experiments on three public Twitter bot detection benchmarks, and results demonstrate that BSG4Bot consistently outperforms all baseline methods, including the recent state-of-the-art methods. In addition, BSG4Bot is trained more efficiently. For example, on the Twitter-22 benchmark, BSG4Bot consumes 23.2% and 21.9% training time compared to the recent related works, RGT [12] and BotMoe [14], respectively. Further experiments also illustrate the effectiveness of different components in BSG4Bot.

## II. Problem Formulation and Data Observation

In this section, we first formulate the problem. We then attempt to uncover content and behavioral features that can potentially differentiate bots from genuine users. We also study the graph homophily in the context of social bot detection, affirming the necessity of considering these factors in the following model design.

### A. Problem Formulation

The social network can be represented as a heterogeneous graph with multi-relations $\mathcal{G} = \{\mathcal{V}, \mathcal{X}, \mathcal{E}, \mathcal{R}\}$, where $\mathcal{V} = \{v_i\}_{i=1}^n$ denotes the set of users, and $\mathcal{X} \in \mathbb{R}^{n \times s}$ represents user features, which have $s$-dimensional vector representations for each node $v$. For any edge relation $r \in \mathcal{R}$, an edge $e_{i,j}^r \in \mathcal{E}_r$ indicates that there is an edge between nodes $v_i$ and $v_j$ under the relation $r$.

With the labeled dataset, the bot detection is to find a function $f : (\mathcal{G}) \longrightarrow Y$ to discriminate whether a node $v_i \in \mathcal{V}$ is a bot or not with the following objectives: $f$ is expected to achieve high performance, in terms of the traditional accuracy and F1 scores. Additionally, the learning of $f$ can be computationally efficient in both time and space cost. Last, $f$ should generalize well to the low training set and unseen node set well.

### B. Observation for Distinguishable Features

The prior works [11], [12], [14] have investigated user features, such as metadata, user descriptions, and tweet contents in identifying bots. As these features are easily imitated or replicated by bots [22], we attempt to find some relatively stable features accumulated over a long-range period.

Our hypothesis posits that the bots tend to exhibit different behaviors from genuine users as the bots are typically invoked to perform specific tasks. From this angle, we report two promising features including Tweet Content Categories and Tweet Temporal Activities, which are verified in the preliminarily in sampled data.

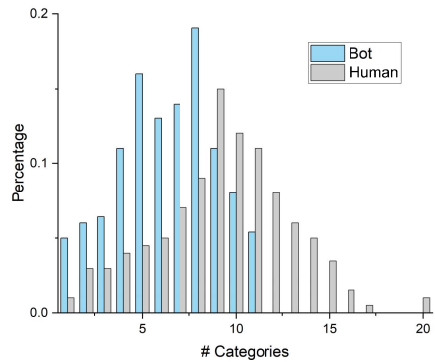

Fig. 2. Distribution of Tweet Content Categories.

**Tweet Content Categories.** We guess that the bots could tend to exhibit narrow focuses in their tweet content categories. In contrast, genuine users, with a broader spectrum of interests and event followings, may display more varieties in their tweet categories. To test this hypothesis, we randomly select 3 communities from the TwiBot-22 [17] benchmark. Each community contains 5,000 bots and 5,000 genuine users. We analyze the content of their last 200 tweets. Using a pre-trained RoBERTa [23] model, we obtain the high-dimensional representation of each tweet. These representations are then clustered into 20 categories using the K-Means algorithm, and the content categories for a user $v$ are defined as the total number of different clusters to which at least one $v$'s tweet belongs.

As illustrated in Figure 2, there is a discernible difference in the distribution of tweet categories between bots and genuine users. In the sampled data, the tweet categories for bots are more focused on specific areas, which could suggest a task-oriented behavior pattern. It is conjectured that this focused distribution may be related to bots being programmed to disseminate certain types of information. In contrast, tweets from genuine users exhibit a broader distribution across various categories, possibly reflecting a more diverse and spontaneous engagement with social media.

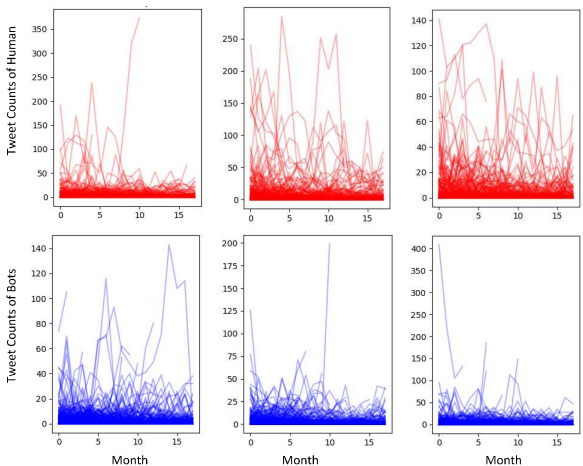

Fig. 3. Counts of tweets posted by users in 3 communities monthly over the past 18 months. Red lines indicate genuine users and blue lines represent bots.

**Tweet Temporal Activities.** We hypothesize that bots and genuine users exhibit different temporal activities features, which is also observed by other works like Spotlight [24] over graph stream. We conjecture that bots are often designed to perform tasks at regular intervals or in response to specific triggers, which may lead to more uniform or predictable patterns of tweet activities. To verify this hypothesis, we randomly select 3 communities and record the number of tweets posted per month by each user over the past 18 months. We plot time series curves of tweet postings for each community to analyze the temporal patterns of tweet activity for both bots and genuine users.

The results, as shown in Figure 3, reveal noticeable differences in tweet activity patterns between genuine users and bots. We can see that genuine users display high variability, dynamic activity spikes, and extremes in tweet counts that are not as prevalent in bots. Bots exhibit more consistent and stable tweeting patterns. These differences support our hypothesis and demonstrate that the temporal characteristics of tweet activities can be leveraged for bot detection.

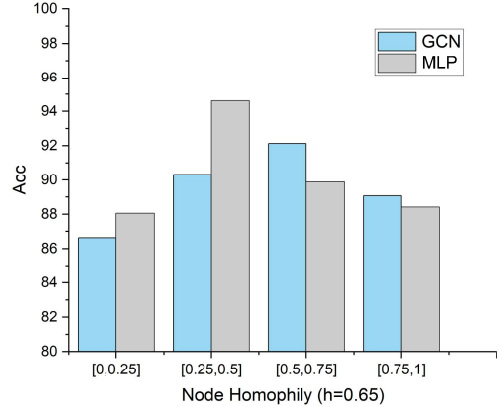

Fig. 4. Relationships between node homophily scores and the accuracy of GCN-based bot detection on MGTAB-22

**Discussion of Features.** We offer two features verified in the sampled data above, provide possible explanation of the features, and will further validate these two features across three datasets in the experimental ablation studies. We stress that the bot detection needs distinguishable features, but does not focus on the detailed differences between two kinds of users. In addition, the bot manipulators may adjust their policy to mimicking the corresponding features of genuine users, even though it takes a long time to change these two features. We will investigate more useful features in the future.

### C. Study on Homophily Ratio

Here, we perform study on the homophily ratio on the given dataset, as the following detection model fully considers the relationship between GNN performance and the node homophily ratio [21], [25]. The node homophily ratio is measured by the average fraction of neighbors with the same labels in Equation 1, where $\mathcal{N}(v_i)$ denotes the neighbor node set of $v_i$ and $d_i = |\mathcal{N}(v_i)|$ is the degree of $v_i$. The center node $v_i$ is considered to be homophilic when more neighbor nodes share the same label as $v_i$ with $h_i > 0.5$.

$$h_i = \frac{|\{u \in \mathcal{N}(v_i) : y_u = y_v\}|}{d_i}, \quad (1)$$

We further define homophily ratio $h$ for an entire graph in the following as the averaged node homophily ratios in Equation 2. A graph is considered homophilic if $h > 0.5$, and heterophilic otherwise. Research has shown that while GNNs perform well in classifying nodes in homophilic graphs where similar nodes are connected, they may underperform in heterophilic settings where connections exist between dissimilar nodes, sometimes even performing worse than simpler models like MLPs [21].

$$h = \frac{\sum_{i \in V} h_i}{|V|} \quad (2)$$

We study the relationship between the GNN (GCN in test) performance on bot detection and node homophily ratio in MGTAB [26] dataset in Figure 4. All nodes are categorized into four groups according to their homophily ratios. The graph has its homophily ratio 0.65, indicating that majority of nodes (more than 65%) falling into the high homophily range (0.5, 1). We also verify the claim [21] that MLP is inferior to GCN on minority nodes (heterophilic nodes in a homophilic graph in our case). For example, MLP achieves better results for nodes with homophily ratios less than 0.5. Such an observation inspires us to increase the homophily ratios of all nodes, *i.e.* by improving the ratios of nodes which shares "similar neighbor pattern" to boost the performance of GCN-based methods in the following.

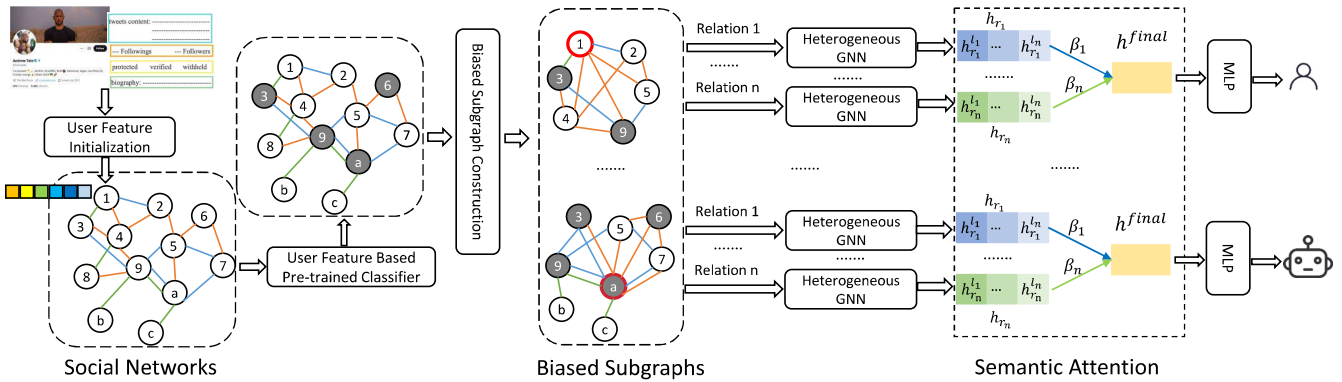

Fig. 5. Architecture of BSG4Bot, Bot Detection based on Biased Heterogeneous Subgraphs.

## III. METHODOLOGY

In this section, we first describe the architecture of BSG4Bot, and then present the major components in detail, including the feature initialization, pre-trained classifier, biased subgraph construction, heterogeneous subgraph learning. We then show the overall training and inference, and analyze the time and space complexity finally.

### A. Framework

We present the architecture of BSG4Bot in Figure 5, illustrated with a toy social network. The entire process can be roughly decomposed into data preparation, subgraph construction, and subgraph learning. In the first phase, the method first extracts user features and user relationships from the social network, and then enriches user features with tweet content categories and temporal activity features discussed above. The combined features are converted into vectors for each node (one vector in Figure 5 for simplicity). Then, we pre-train a coarse classifier using efficient multilayer perception (MLP) model over all nodes in the graph. We can see that the nodes are roughly classified. For example, nodes such as 3, 9 with the gray color have more chances to be bots, while other nodes are likely be genuine users.

The subgraph construction is a key step in BSG4Bot. For each node $v$ in the graph, we construct the subgraph starting from $v$ (red circle in Figure 5), in which the structural importance as well as the node homophily ratio are considered. The subgraph is termed biased, as the neighbors nodes sharing the same label to $v$ have more chances to be selected into the subgraph. For example, we can see more gray nodes are added into the subgraph starting from node $a$, which is also a gray node.

The subgraph learning is the final step in BSG4Bot. As different relationships may exist in the sampled graph, we adopt the idea of RGCN [27] to learn patterns from heterogeneous subgraphs. That is, we extract multiple homogeneous graphs each with one relationship, and apply semantic attention to combine these different graphs. The hidden states in different layers are concatenated between consolidation, as different layers capture different extents of homophilic features.

### B. Node Feature Initialization

Node features for the following pre-trained classifier and GNN model are initialized into $x_i$ as follows. Here, $x_{d,i}$, $x_{t,i}$, $x_{p,i}^{\text{num}}$, $x_{p,i}^{\text{cat}}$, $x_{t,i}^{\text{cate}}$, and $x_{t,i}^{\text{time}}$ represent the user description, tweet content, numerical metadata features, categorical properties of metadata, tweet categories, and tweet temporal activities, respectively. Among them, $x_{d,i}$, $x_{t,i}$, $x_{p,i}^{\text{num}}$, and $x_{p,i}^{\text{cat}}$ are extracted similarly to those in BotRGCN [11].

$$x_i = \left[ x_{d,i}; x_{t,i}; x_{p,i}^{\text{num}}; x_{p,i}^{\text{cat}}; x_{t,i}^{\text{cate}}; x_{t,i}^{\text{time}} \right] \qquad (3)$$

Two features discussed in Section 3 are also extracted and encoded into the user features. For the content category feature $x_{t,i}^{\text{cate}}$, we select the most recent 200 tweets for each user. These tweets are encoded with RoBERTa and then clustered into 20 categories using the K-means algorithm. The number of tweet categories for each user is normalized using z-score normalization. Additionally, we calculate the percentage of tweets in each category for each user. The z-score normalized number of tweet categories is concatenated with the percentage of tweets in each category. For the user's tweet categories, the concatenated result is processed through another fully connected layer, yielding $x_{t,i}^{\text{cate}}$.

For the temporal activity feature $x_{t,i}^{\text{time}}$, we first extracted the number of tweets posted by each user in the past 12

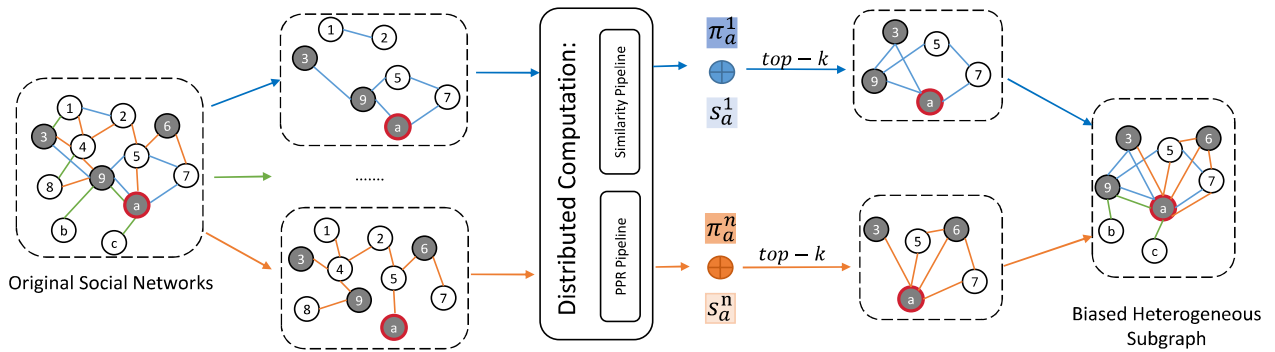

Fig. 6. An illustration of Biased Heterogeneous Subgraph Construction Rooted at Node **a**.

months. To handle accounts with fewer tweets and ensure feature alignment, we represent months without tweets by filling them with zeros. The percentage of tweets posted each month is then calculated to form the temporal activity features of the user. This temporal activity data is passed through a fully connected layer to obtain the feature representation of the user's tweet temporal activities, denoted as $x_{t,i}^{\text{time}}$.

### C. Pre-trained Classifier on User Features

Initially, we introduce a pre-trained model to assist in the following biased subgraph construction. We use the MLP on user features in the pre-classification method, as the MLP model is a simple yet efficient method to achieve sufficient precision, e.g. with F1-score 81% in Twitter-20, or with F1-score 53% in a more complex benchmark Twitter-22. In this way, an MLP model can be used as an effective tool to enhance the node homophily by selecting the neighbors with the same label.

Specifically, we train a two-layer MLP on both the training and validation sets to preliminarily obtain the probability that a user is a human or a bot by leveraging cross-entropy loss in Equation 4, where $W_0$, $W_1$, $b_0$, and $b_1$ are learnable parameters. $\sigma$ represents an activate function and we adopt leaky-relu as $\sigma$ for the rest of the paper.

$$\hat{Y} = \text{softmax}(\sigma(W_0 \cdot X + b_0)W_1 + b_1) \quad (4)$$

Subsequently, we obtain the hidden representations for users extracted by the pre-trained MLP model, and then calculate the similarity between the starting node $v_i$ and a neighbor node $v_j$ based on the output of the pre-trained MLP in Eq. 6, where $cos(.,.)$ represents the cosine similarity. The cosine similarity score $s_{i,j}$ is normalized to the range [0,1]. This similarity reflects the proximity between user nodes in the feature space and influences the sampling probability of each neighbor node when constructing the biased subgraph.

$$h_i^{mlp} = W_0 \cdot x_i + b_0 \quad (5)$$

$$s_{i,j} = \frac{1 + \cos(h_i^{mlp}, h_j^{mlp})}{2} \quad (6)$$

### D. Biased Subgraph Construction

The subgraph construction is the key component in the BSG4Bot, which aims to achieve two important goals in the bot detection. For each node $v$, the neighbor nodes have more chances to be selected into the $v$'s subgraph if they are likely to share the same label as $v$, and thus homophilic ratios are expected to increase, which further improve the performance of a GNN model. The model is trained using batches of subgraphs rather than the entire graph, so as to explore the available computational resources in a flexible way.

We consider three factors, including heterogeneous edge relations, node importance in graphs, and node homophily in the subgraph construction. We use the graph in Figure 5 to explain the detailed subgraph construction steps in Figure 6. We first extract homogeneous graphs for each edge relationship separately from a heterogeneous graph. For each starting node $v$, we then select top-$k$ neighbors into $v's$ subgraph by combining node importances using PPR and node homophily ratio using the pretrained classifier. Note that the selected neighbors are not restricted by the direct neighbors, but the nodes which may play important roles in prediction on $v$, no matter hops to the $v$. The distributed version of PPR and similarity computation are used to reduce the cost in the subgraphs. Finally, we combine these subgraphs to form a heterogeneous subgraph with multiple edge relations, from which the bot detection patterns are learned. In the following, we discuss these steps in detail.

**Heterogeneous Edge Relations.** We consider the heterogeneous edge relations in the social network. In social networks, users interact through "following" and "follower" relationships, and also communicate through tweets such as "mention" other users in a tweet, "reply to" someone's tweet, or "like" a tweet. The "following", "follower", "mention" and other relationships between users have different impacts on bot detection [12]. In order to capture these impacts of varied relations, we build multiple subgraphs starting from the same starting node, each containing the same type of relation only.

**Node Importance in Graphs.** We then take the node importance into the subgraph construction. Usually, the node importance can be measured by the PPR score [28]. In social networks, while calculating the PPR score from a particular user, nodes with higher scores may represent loyal followers or users who frequently interact with that user.

The PPR algorithm modifies traditional PageRank by adding a restart option to a specific starting node during the random walk. At each step, there's a chance that the walk will return to the starting node, thereby reflecting their importance to the starting node. For any starting node $v_i$, the calculation of its PPR score vector $\pi_i$ can be formally expressed as Equation 7, where $\alpha \in (0, 1)$ is a predefined parameter called the teleportation probability, and the indicator vector $e_i$ is called the preference vector for defining PPR.

$$\pi_i = \alpha(I - (1 - \alpha)D^{-1}A)^{-1}e_i \tag{7}$$

We utilize an approximate method [29] to efficiently compute the PPR score. Roughly speaking, we initialize the residual score to be 1 for the starting node $v$ and 0 for other nodes. According to the teleportation probability, part of residual scores are kept at the local nodes, and the remaining scores are distributed to the neighbor nodes. As the current node may also receive residual scores from its neighbors, the newly-added residual scores are actually distributed recursively, until the newly-added residuals are sufficiently small. Then, the residual scores on other nodes can serve as the importances to $v$.

**Node Homophily.** The subgraph construction should also consider the node homophily, which plays a role complementary to the node importances computed by PPR. Then, we define a combined score in Equation 8, where $\pi_{ij}$ denotes the PPR score of the neighbor node $v_j$ with respect to the starting node $v_i$, and $s_{ij}$ represents the similarity between the starting node $v_i$ and its neighbor $v_j$ calculated by Equation 6 using the pre-trained classifier. BSG4Bot assumes that both PPR scores and node homophily ratio are considered equally important, hence $\lambda$ is set to 0.5.

$$p_{ij} = \lambda \pi_{ij} + (1 - \lambda)s_{ij} \tag{8}$$

Algorithm 1 outlines the overall process for creating biased heterogeneous subgraphs in BSG4Bot starting from a node $v$. From Line 1 to Line 7, different homogeneous graphs are constructed for each kind of edge relationship, in which the PPR scores are first computed as $\pi^r(v)$ in Line 3. The scores used in the subgraph construction in Line 5 combine node similarity scores from the pre-defined classifier and the node importance scores, from which the top-$k$ nodes are selected into the subgraph.

The edge in the subgraph can be constructed as follows: All the selected nodes establish links to the start node $v$ no matter whether there is an edge between them in the original graph. In addition, the edges from the original graph are retained in the corresponding subgraphs. Thus, each constructed subgraphs is connected, which facilitates the features aggregation in the following GNN training.

---

**Algorithm 1: Biased Subgraph Construction**

**Input:** Multi-relations Graph $\mathcal{G} = (V, E, R)$,
Pre-trained representation $\hat{Y}$, start node $v$, $k$.
**Output:** Biased Subgraphs $\mathcal{G}_v$.

1 **for** *relation $r$ in $R$* **do**
2     Extract graph $\mathcal{G}^r$ under relation $r$: $\mathcal{G} = (V, E, r)$;
3     Compute PPR vector $\pi^r(v)$ and locate PPR neighbors $N^r(v)$ in $\mathcal{G}^r$;
4     Compute node similarity
      $s = (1 + \cos(\hat{y}_v, \hat{y}_{N^r(v)}))/2$ using $\hat{Y}$;
5     Compute combined score vector $p = \pi^r(v) + s$;
6     Select nodes into $N_k^r(v)$ with top-$k$ combined scores from $p$;
7 Initialize $\mathcal{G}_v$;
8 **for** *relation $r$ in $R$* **do**
9     **for** *node $v_j$ in $N_k^r(v)$* **do**
10        **for** *node $v_k$ in $N_k^r(v)$* **do**
11           **if** *edge $e = (v_j, v_k, r)$ in $E$* **then**
12              Add $e$ to $G_r$;
13           Add $(v_j, v_k, r)$ to $G_r$ if $v_j = v$;
14     $\mathcal{G}_v \leftarrow \mathcal{G}_v \cup \{G_r\}$
15 **return** $\mathcal{G}_v$;

---

### E. Heterogeneous Subgraph Learning

Similar to the RGCN, we utilize GNN models on subgraphs generated for each relation type to obtain the hidden representation of the starting node, and then apply semantic attention layers to combine the representation from different graphs. We further consider concatenating the hidden representation from different layers, as they carry different information to overcome the possible mixed pattern in the subgraphs.

**Graph Encoder.** We first transform the user features to obtain hidden vectors as Equation 9, where $W_2$ and $b_2$ are learnable parameters.

$$h_i^0 = \sigma(W_2 \cdot x_i + b_2) \tag{9}$$

Then, we utilize a GCN [30] for each subgraph with the same edge relation to learn the node embeddings. At the $l$-th layer, the representation of a node under the relation $r$ is defined as Equation 10, where $c_i$ represents a normalization constant, and $N_i$ denotes the one-hop neighbors of node $v_i$.

$$h_i^l(r) = \sigma\left(\sum_{j \in N_i} \frac{1}{c_i} W_3^{(l)} h_j(r)^{(l-1)}\right) \tag{10}$$

**Intermediate Representation Concatenation.** There may still exist a mixture of homophily and heterophily in subgraphs even though we have taken the biased subgraph construction strategy. In theory, a GCN layer can be viewed as a low-pass filter [31], where intermediate outputs of the shallower layers contain higher-frequency components than that in the deeper layers. From the perspective of graph homophily, minority-class nodes tend to exhibit more information in the high-frequency components [32].

Inspired by these insights, we concatenate the intermediate outputs of different GNN layers, which can effectively capture information that characterizes different frequencies. The final node representation at the $l$-th layer in relation $r$ can be formulated as Equation 11, where $h_i^{l_{final}} \in \mathbb{R}^{(l+1)s}$.

$$h_i^{l_{final}}(r) = \text{COMBINE}\left(h_i^0(r), \ldots, h_i^l(r)\right) \quad (11)$$

**Semantic Attention Layer.** The representation of each node in heterogeneous graphs with multiple relations is obtained through multiple GCN-based embedding layers. Considering the varying significance of relations [33], we employ a semantic attention layer to integrate the representations across different relations. The importance of each relation, denoted as $w_r$, is shown as Equation 12, where $W$ is a weight matrix, $b$ is a bias vector, and $q$ is a semantic-level attention vector. We have all the parameters mentioned above shared across all relations and semantic-specific embeddings in BSG4Bot.

$$w_r = \frac{1}{|\mathcal{V}|} \sum_{i \in \mathcal{V}} q^{\text{T}} \cdot \tanh\left(W \cdot h_i^{l_{final}}(r) + b\right), \quad (12)$$

Then, we normalize the importances of all relations using a softmax function. The weight of relation $r$, denoted as $\beta_r$, can be obtained by normalizing the above importances of all relations using the softmax function in Equation 13, which can be interpreted as the contribution of the relation $r$ to a specific task. The higher $\beta_r$, the more important the relation $r$ is. Notably, for different heterogeneous graphs, relation $r$ may have different weights.

$$\beta_r = \frac{\exp(w_r)}{\sum_{r=1}^{R} \exp(w_r)}, \quad (13)$$

With the learned weights as coefficients, we can fuse these semantic-specific embeddings to obtain the final embedding $h_i^{final}$ as follows:

$$h_i^{final} = \sum_{r=1}^{R} \beta_r \cdot h_i^{l_{final}}(r) \quad (14)$$

*F. Model Training*

We employ a softmax layer to make predictions on the final user representations from the graph neural network, using Equation 15, where $\hat{y}_i$ is the prediction of user $v_i$.

$$\hat{y}_i = \text{softmax}\left(W_O \cdot h_i^{final} + b_O\right) \quad (15)$$

The loss function of BSG4Bot is constructed using Equation 16, where $y_i$ represents the ground-truth label, and $\theta$ encompasses all learnable model parameters.

$$L = -\sum_{i \in Y} [y_i \log(\hat{y}_i) + (1 - y_i) \log(1 - \hat{y}_i)] + \lambda \sum_{w \in \theta} w^2 \quad (16)$$

The training can be performed in a batch manner. That is, for each node in the training set, we perform the subgraph construction, and store the constructed subgraphs. Then, during each training epoch, we compose a training batch from sampled subgraphs, which requires much lower computational resources than that on the entire graph in the training.

*G. Complexity Analysis*

We first introduce the symbols used in the following analysis. Let $n$ be the total number of nodes, $f$ be dimension of the features, $d$ be the average degree of nodes, $k$ be the total number of nodes in the subgraphs, $h$ be the dimension of the hidden state, and $\ell$ be the layers in the neural network. For simplicity, we assume that both MLP and GNN networks share the same $h$ and $\ell$.

We analyze the time and space complexity of our proposed BSG4Bot. The following analysis is divided into three main components: pre-trained MLP classification, biased subgraph construction, and heterogeneous subgraph learning, as we may need different epochs for training in these components.

It involves forward and backward propagation in training an MLP classifier. For each training epoch, it costs time $O(n \cdot f \cdot h \cdot \ell)$, as the MLP operates on features of all users separately. The space cost is dominated by $O(n \cdot f)$, and the space for the parameters is much less than that for the graph node features.

In the second phase of biased subgraph construction, BSG4Bot computes the PPR scores in time complexity $O(n \cdot d \cdot \log d)$, when the approximate method is used. The combined similarity scores need the node similarities using the pre-trained MLP classifier, which takes $O(n \cdot k \cdot h)$, when using the approximate PPR scores to limit the candidate nodes to be compared. The location of the nodes with the top-$k$ combined similarities can be computed in $O(n \cdot k \cdot \log k)$ to sort the candidate nodes. The total time complexity is dominated by the PPR calculation, resulting in $O(n \cdot d \cdot \log d)$. The space complexity in this step is $O(|G|)$, which is needed to store the node features and relationships in the main memory.

The subgraph learning processing takes $O(b \cdot k \cdot h \cdot \ell \cdot n/b)$ for each epoch, as each node in a batch with $b$ nodes along with its $k$ neighbors propagate their features into neighbors in one layer, which are recursively processed in the following $\ell$ layers. By considering the total number nodes in one epoch, the time cost becomes $O(n \cdot k \cdot h \cdot \ell)$. The space complexity takes $O(b \cdot k \cdot h \cdot \ell)$, in which only the batched sample subgraphs are loaded into the memory for training.

Although both full-graph GNN training and BSG4Bot need to load the entire graph, including features, into the memory, the GNN model training needs GPU memory, while BSG4Bot can compute the PPR scores in the main memory, and process much smaller subgraphs $O(b \cdot k)$ than the entire graph $O(n)$ during the model training, which greatly improves the scalability. In addition, the enhanced homophily in the subgraphs can significantly lower the number of the epochs before model convergence, which results in BSG4Bot's efficiency.

## IV. EXPERIMENT

In this paper, we first present the experimental settings and compare BSG4Bot to other methods. Then, we focus on the roles of subgraphs, and study the effects of different components in the BSG4Bot.

### A. Experimental Setup

**Datasets.** We evaluate BSG4Bot on three widely-adopted Twitter bot detection benchmarks, including TwiBot-20 [34], TwiBot-22 [17] and MGTAB [26]. Table I summarizes the statistics of each benchmark. We follow the original train, validation, and test splits of the benchmarks for a fair comparison to previous works. We should note that *TwiBot-22* dataset provides additional 10 non-overlapped communities, each of which contains 10,000 nodes, with 5,000 labeled as bot and 5,000 labeled as genuine users. These communities can be used to validate the generalization of the detection models.

TABLE I
STATISTICS OF BENCHMARKS.

| Benchmarks | TwiBot-20 | TwiBot-22 | MGTAB |
|---|---|---|---|
| # users | 229,580 | 1,000,000 | 10,199 |
| # human | 5,237 | 860,057 | 7,451 |
| # bot | 6,589 | 139,943 | 2,748 |
| # edges | 227,979 | 3,743,634 | 1,700,108 |
| # relations | 2 | 2 | 7 |

**Baselines.** We compare BSG4Bot to the following methods roughly in 5 categories, including basic methods (1-2), traditional GNN models (3-4), GNN models with sampler (5-7), the existing bot detection methods (8-10), and the GNN models that consider homophily (11-12).

1) **RoBERTa** [23] encodes user descriptions and tweets using pre-trained RoBERTa, and feeds user features into an MLP for bot identification.
2) **MLP** here is actually the pre-classifier in BSG4Bot.
3) **GCN** [30] aggregates weighted features from neighbors, and passes user representations into an MLP for classification.
4) **GAT** [35] introduces the attention mechanism to distinguish the importances of neighboring users in aggregation, before feeding into an MLP for classification.
5) **SlimG** [36] achieves efficient training on large-scale graph data through a simplified model architecture, hyperparameter-free propagation functions, and effective preprocessing of features.
6) **GraphSAGE** [37] performs uniform sampling in collecting neighbors in aggregation, which are then passed into an MLP for bot detection.
7) **ClusterGCN** [18] conducts GNN training over the subgraphs, each of which is a combined result of different clusters, thus enhancing the scalability of GNN training.
8) **BotRGCN** [11] constructs a heterogeneous graph on Twitter and exploits relational graph convolution networks for user representation learning.
9) **RGT** [12] employs relational graph transformers to leverage relation and influence heterogeneity of the Twitter networks.
10) **BotMoe** [14] adopts a community-aware mixture-of-experts architecture to learn various patterns in different communities.
11) **H2GCN** [32] identifies a set of key designs that can boost learning from a heterophilic graph without trading off accuracy in homophilic structure.
12) **GPR-GNN** [38] learns to jointly optimize node feature and topological information extraction adaptively, regardless of the extent to which the nodes are homophilic or heterophilic.

**Implementation.** BSG4Bot is implemented in PyTorch, PyTorch Geometric, Transformers, Scikit-learn, and NumPy. All our experiments are performed on a server with 256GB RAM, two Intel (R) Xeon (R) Silver 4210R CPUs @ 2.40GHz and a 24GB GeForce RTX 3090 GPU. To avoid overfitting, dropout and early stopping techniques are used for training.

### B. Performance and comparison

In this subsection, we compare BSG4Bot to various baseline models with the same training and test set in terms of the accuracy and F1-score metrics to study effectiveness. The performance with varying training sets is also studied. We then examine the training efficiency in terms of the training time and convergence rates. Finally, we study the generalization to unseen data of different methods.

**Performance on different Baselines.** We compare BSG4Bot with 12 baseline methods. The performance is measured in terms of Accuracy (Acc) and F1-score. We run each experiment for 5 times with random weight initializations and report the average value as well as the standard deviation on the test set.

The experimental results are shown in Table II. We can see that BSG4Bot outperforms all baseline methods across all three benchmarks. Specifically, BSG4Bot demonstrates enhancements over the state-of-the-art Twitter bot detection method BotMoe, achieving improvements of 1.0% in accuracy and a notable **4.5%** in F1-score on TwiBot-22. Similarly, on the MGTAB dataset, BSG4Bot outperforms RGT with gains of 2.5% in accuracy and 2.3% in F1-score. Furthermore, on Twibot-20, BSG4Bot outperforms BotMoe by 1.5% in accuracy and 0.6% in F1-score.

In addition, the experiments demonstrate that the graph homophily factor should be considered in the bot detection. As shown in Table II, a simple MLP can outperform GCN on all three benchmarks, outperform GAT on TwiBot-20 and MGTAB, and outperform GraphSAGE on TwiBot-22. At the same time, H2GCN, GPR-GNN and BSG4Bot (ours), which is designed to consider heterophilic factors in graphs, outperform

TABLE II
ACCURACY AND F1-SCORE OF COMPETITORS ON THE THREE BENCHMARKS.

| Model | Twibot-20 | | Twibot-22 | | MGTAB | |
|---|---|---|---|---|---|---|
| | Accuracy | F1-score | Accuracy | F1-score | Accuracy | F1-score |
| RoBERTa | 75.5(0.1) | 73.18(0.5) | 72.12(0.1) | 20.51(1.5) | - | - |
| MLP | 83.89(1.1) | 81.71(0.8) | 79.01(0.7) | 53.81(0.3) | 84.88(0.4) | 84.67(0.3) |
| GCN | 77.52(0.2) | 80.85(0.4) | 78.41(0.4) | 54.91(0.4) | 83.65(0.2) | 84.02(0.6) |
| GAT | 83.33(0.3) | 81.26(0.7) | 79.54(0.3) | 55.83(0.9) | 84.45(0.3) | 83.69(0.4) |
| GraphSAGE | 84.57(0.5) | 84.57(0.5) | 76.74(0.4) | 45.44(2.1) | 86.72(0.6) | 84.95(1.1) |
| ClusterGCN | 85.23(0.4) | 85.36(0.3) | 78.45(0.2) | 56.87(1.8) | 88.73(0.5) | 85.35(0.6) |
| SlimG | 86.55(0.3) | 87.97(0.5) | 74.76(0.3) | 44.27(1.6) | 88.13(0.3) | 84.45(0.7) |
| BotRGCN | 85.86(0.8) | 87.33(0.7) | 78.56(0.1) | 57.52(1.2) | 89.69(1.1) | 86.02(1.2) |
| RGT | 86.67(0.3) | 88.22(0.1) | 76.44(0.2) | 43.02(0.7) | 89.76(0.4) | 86.59(0.8) |
| BotMoe | 87.84(0.4) | 89.32(0.3) | 79.16(0.1) | 56.87(0.3) | - | - |
| H2GCN | 88.23(0.6) | 89.14(0.7) | 77.64(0.4) | 57.23(1.3) | 90.56(0.5) | 87.72(0.3) |
| GPR-GNN | 87.47(0.8) | 88.84(1.2) | 78.64(0.6) | 57.66(0.8) | 90.32(0.4) | 87.46(1.1) |
| BSG4Bot (Ours) | **89.15(0.4)** | **89.89(0.2)** | **79.93(0.2)** | **59.42(1.3)** | **92.25(0.7)** | **88.92(1.5)** |

TABLE III
COMPARISONS OF RUNNING TIME ON TWIBOT-22 BENCHMARK.

| | Time per Epoch | #Epochs | Total training time |
|---|---|---|---|
| GCN | 4min17s | 165 | 11h45min |
| GAT | 4min42s | 176 | 13h47min |
| GrapgSAGE | 4min47s | 178 | 14h11min |
| ClusterGCN | 4min10s | 76 | 5h16min |
| SlimG | **2min16s** | **62** | **2h21min** |
| BotRGCN | 4min38s | 163 | 12h35min |
| RGT | 6min36s | 192 | 21h07min |
| BotMoe | 7min06s | 187 | 22h08min |
| H2GCN | 5min04s | 172 | 14h31min |
| GPR-GNN | 5min16s | 169 | 14h50min |
| BSG4Bot(ours) | 4min22s | 67 | 4h52min |

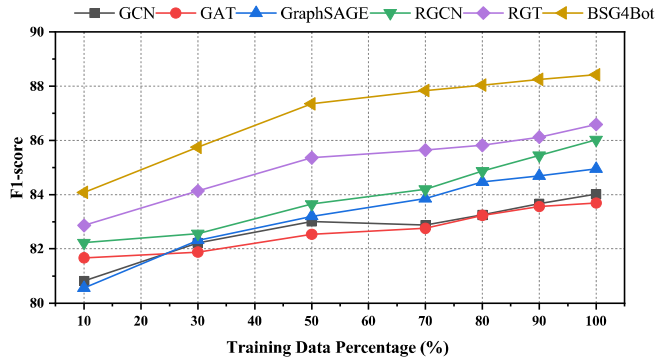

Fig. 7. Performance of Competitors Varying Percentages of Labeled Users on MGTAB.

other graph-based methods, highlighting the challenge of mixed patterns on social bot detection.

**Runtime and Convergence.** We record the average training time per epoch, to investigate the running time of GNN based competitors. Table III reports the time per epoch and total number of epoches used, where "# Epochs" refers to the number of training epochs before early stopping is triggered due to a lack of improvement on the validation set. We can see that BSG4Bot has an overwhelming advantage in terms of total training time, requiring only 62 epochs to reach early stopping while time cost per epoch is similar. BSG4Bot requires only 23.2% training time compared to RGT, and 21.9% training time compared to BotMoe. Such a performance gains come from the relatively easy training process on biased subgraphs with the enhanced homophily. SlimG is considerably faster than BSG4Bot when it comes to runtime, completing the task on the Twibot-22 dataset in a little over half the time. However, SlimG is not effective in the bot detection, as its accuracy is reduced by 6.47%, and its F1 score is reduced by 25.50% compared to BSG4Bot, as illustrated in Table II.

**Performance with Low Samples.** The learned bot detection methods rely on the labeled data, while labeling a bot requires a domain expert to check the content and structural features carefully, and even to perform offline investigation in some cases. Thus, the methods that work well with low samples are highly needed for the bot detection. Here, we conduct experiments varying proportions of labeled users to assess the sample efficiency of different models. The proportions are incrementally increased from 10% to 100% of training data, providing insights into how model performance scales with the amount of available annotated data.

Figure 7 illustrates that BSG4Bot (indicated by the purple line) consistently outperforms other methods, demonstrating high F1-score even with low sample sizes. The relative advantages of our methods over other methods are stable, ranging from 1% to 2% in terms of absolute scores of F1. Specifically, the F1-scores of BSG4Bot decreases from 89% on full training data to 84% on only 10% training data, making BSG4Bot suitable for practical low-resource applications. Similar results can be observed on TwiBot-20 and TwiBot-22, but are omitted here due to space limitations.

**Generalization Study.** The generalization of a model is also required by the detection methods, as the bots constantly

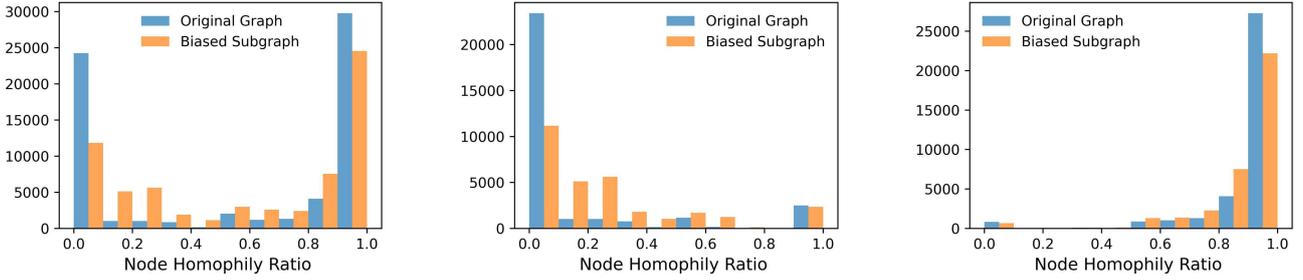

(a) All Users. Average node homophily rate of all users increased from $h_{\text{avg}} = 0.585$ in the original graph to $h_{\text{avg}} = 0.610$ in the biased subgraphs.

(b) Bots. Average node homophily rate of bots increased from $h_{\text{avg}} = 0.127$ in the original graph to $h_{\text{avg}} = 0.180$ in the biased subgraphs.

(c) Human. Average node homophily rate of human decreased from $h_{\text{avg}} = 0.975$ in the original graph to $h_{\text{avg}} = 0.973$ in the biased subgraphs.

Fig. 8. Node Homophily Ratio Distributions of The Original Graph and The Biased Subgraphs on Twibot-22 Benchmark.

evolve in the battle between bot manipulators and bot detection, and the model had better to detect unseen bots. Recall that TwiBot-22 benchmark provides 10 non-overlapping communities. We select four bot detection models (including BSG4Bot), train these models on one community, and apply the trained models to the remaining 9 communities, so as to study generalization ability on unseen communities within the TwiBot-22 benchmark.

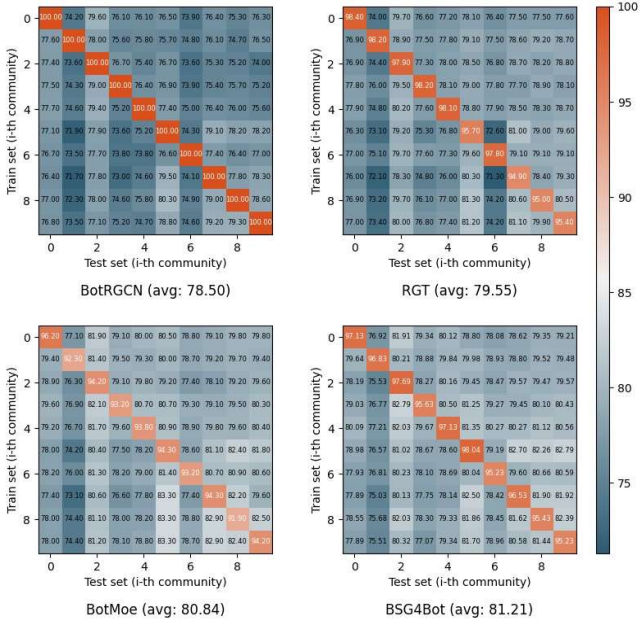

Fig. 9. Performance of Competitors on Unseen Communities.

Figure 9 reports the model performance on the $i$-th community when the model is trained on the $j$-th community. We can see that BSG4Bot is better at generalizing to unseen communities. BSG4Bot achieves the highest average accuracy of 81.21 among all bot detection approaches. The gain in the generalization, we think, comes from the extracted features shared among different nodes and the effective model used in BSG4Bot.

### C. Biased Subgraph Study

Now, we go deeper into the biased subgraph to study whether the homophily ratios really increase in the biased subgraph constructed, the effects of the size of subgraph on the performance, and whether the biased subgraph constructed can be a valid plugin to other GNN models.

**Study of Subgraph Homophily Ratio.** We have claimed that the increase of the homophily ratios can lead to the increase of the GNN performances, and BSG4Bot has illustrated performance advantages above. Now, we are interested in whether the homophily ratio really increases in the biased subgraph compared to that in the original graph.

Figure 8 presents the distribution of node homophily ratios for all users, bots, and genuine users on Twibot-22 benchmark, respectively. As expected, we can see that the homophily ratios in the biased subgraphs are generally higher than those in the original graph for all users and bots. More specifically, the improvement is particularly notable for bots, suggesting that BSG4Bot excels at aggregating bot accounts with similar characteristics. It is a crucial factor to improve the capability in determining a bot. We also see a slight decrease of homophily ratios (but still near 1) for genuine users. It is due to that the PPR scores are also considered in the subgraph construction.

**Study of Subgraph Size.** We test the accuracy and F1-score of bot detection varying $k$ to investigate the impact of the biased subgraph size $k$ on model performance. As shown in Figure 10, we observe that when the subgraph size is relatively small, an enlarged subgraph leads to improvements in both accuracy and F1-score. This suggests that including more neighbors, up to a point, is needed in BSG4Bot, as these neighbors have high chances to share the same label as the starting node, and contribute positively as the high-order features. However, as $k$ further increases (64 to 128 in Twibot-20 and Twibot-22, 16 to 128 in MGTAB), we observe a slight decrease in performance. It is partially due to that beyond a certain threshold, it is inevitable to select heterophilic nodes into the subgraph, and such a mixture heterophilic and homophilic pattern is still a challenge for the current GNN models.

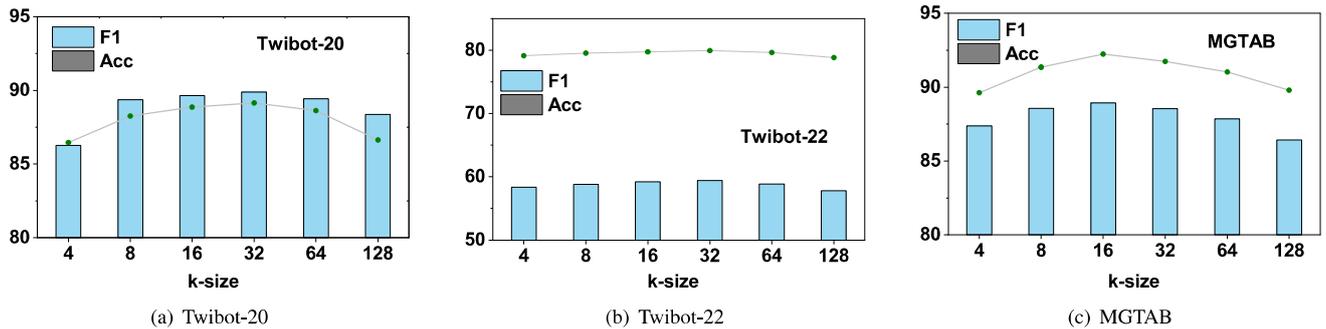

Fig. 10. Performance of BSG4Bot across Various Subgraph Size.

**Study of Biased Subgraph as a General Plugin Component.** BSG4Bot trains a GNN model over the biased subgraph with the enhanced homophily ratios. We are interested in whether the subgraphs constructed can serve as a general plugin component used before other GNN models. In order to do so, we integrate the biased subgraphs into GCN, GAT, and BotRGCN models, and test their accuracy and F1 scores across three datasets.

Table IV shows that a significant improvement across all the three models when our biased subgraph is incorporated, which further verifies the effectiveness of the biased subgraph. In addition, it is reasonably expected that the biased subgraph plugin may be used in other GNN-based downstream tasks, which will be further studied in the future.

TABLE IV
ACCURACY AND F1-SCORE ON THE THREE BENCHMARKS OF BIASED SUBGRAPHS AS A PLUG-AND-PLAY COMPONENT ON GNNS.

| Model | Twibot-20 | | Twibot-22 | | MGTAB | |
|---|---|---|---|---|---|---|
| | Acc | F1 | Acc | F1 | Acc | F1 |
| GCN | 77.52 | 80.85 | 78.41 | 54.91 | 83.65 | 84.02 |
| Subgraphs + GCN | 84.64 | 86.08 | 78.63 | 55.68 | 84.96 | 85.31 |
| GAT | 83.33 | 81.26 | 79.54 | 55.83 | 84.45 | 83.69 |
| Subgraphs + GAT | 85.15 | 86.69 | 79.55 | 56.34 | 86.47 | 85.11 |
| BotRGCN | 85.86 | 87.33 | 78.56 | 57.52 | 89.69 | 86.02 |
| Subgraphs + BotRGCN | 86.81 | 88.19 | 79.01 | 57.84 | 90.39 | 86.65 |
| BSG4Bot (Ours) | **89.15** | **89.89** | **79.93** | **59.42** | **92.25** | **88.92** |

*D. Ablation Study*

We conduct ablation studies on three benchmarks to highlight the contributions of different components to the overall performance of our framework, including the subgraph construction rules, the two features introduced in Section II, and the concatenation of intermediate results and semantic attention in aggregation. The results are shown in Table V.

We can draw the following conclusions:
- Overall results. These results demonstrate that each component of the model contributes to the overall performance of BSG4Bot. We can observe that replacing or removing any component of BSG4Bot results in performance degradation.
- Without the tweet category feature or the temporal activity burst feature. We can see that omitting these features leads to a performance drop on both benchmarks (the results of removing temporal features are not included on Twibot-20 due to the lack of tweet time in raw data). The results demonstrate both features can be included in BSG4Bot as they can be efficiently collected, and enhance the model's performance at a low cost.
- Without intermediate representation concatenation. The absence of concatenation of intermediate representations lowers performance on both benchmarks, which confirms the effectiveness of leveraging multiple layers' representations in GNN to handle the remaining mixture of homophily and heterophily in subgraphs.
- Replacing biased subgraphs with PPR subgraphs. The biased subgraph is the most efficient strategy in BSG4Bot. We can observe that a noticeable decline in both accuracy and F1-score on two benchmarks. It is due to that BSG4Bot enhances the homophily of subgraphs, especially on Twibot-20 benchmark, where the pre-classifier achieves high precision.
- Replacing semantic attention with mean pooling. The semantic attention layer is crucial for BSG4Bot. When replacing it with mean pooling, there is a significant drop in both accuracy and F1-score across all benchmarks. This is because the semantic attention layer effectively integrates node representations by prioritizing more important relations, enhancing feature representation and reducing noise. The mean pooling approach fails to account for the varying importance of relations, leading to less effective node embeddings and overall reduced performance.

## V. RELATED WORK

In this section, we review the advances of the social bot detection and further discuss two extensions to the GNN models related to this paper, including heterophilic graph learning, and graph sampling methods.

TABLE V
ABLATION STUDY OF BSG4BOT ON THREE BENCHMARKS.

| Ablation Settings | Twibot-20 Acc | F1 | Twibot-22 Acc | F1 | MGTAB Acc | F1 |
|---|---|---|---|---|---|---|
| full model | 89.15 | 89.89 | 79.93 | 59.42 | 92.25 | 88.92 |
| w/o tweet category feature | 88.56 | 89.24 | 79.47 | 59.36 | - | - |
| w/o tweet temporal feature | - | - | 79.54 | 59.23 | - | - |
| replacing biased subgraphs with PPR subgraphs | 87.35 | 87.04 | 79.07 | 58.33 | 89.93 | 86.76 |
| w/o intermediate representation concatenation | 88.16 | 89.03 | 79.26 | 58.86 | 91.54 | 87.92 |
| replacing semantic attention with mean pooling | 87.98 | 88.49 | 79.36 | 58.74 | 91.36 | 87.85 |

*A. Social Bot Detection*

**Feature-based methods** primarily extract features from user metadata [39], descriptions [8], temporal patterns [40], tweets [41], and sentiment features [42], and then employ traditional machine learning techniques such as Random Forests, SVMs, and K-means for bot detection. However, the effectiveness of these methods may be compromised when bots are engineered with complex feature manipulations [43].

**Content-based methods** primarily encode textual information from user descriptions and tweets using deep neural networks. NLP techniques such as LSTM and BERT are employed in bot detection. Wei et al. [44] use a combination of Bi-directional Long Short-Term Memory (BiLSTM) models and word embedding techniques. Another study [45] employs BERT for sentiment classification of user tweets, extracting emotional features to aid in bot detection. BGSRD [46] constructs a large text graph containing word and tweet nodes, which are then fed into a GCN [30]. However, the performance of content-based methods faces challenges when advanced bots mimic features of genuine users [22], such as replication or transformation of tweets and descriptions.

**Graph-based methods** have emerged as a significant area of interest in recent research. These methods usually adopt or extend the existing graph neural networks in detecting bots. Alhosseini et al. [10] pioneer the use of graph neural networks in bot detection tasks. BotRGCN [11] is proposed to construct heterogeneous graphs with varying relationships on Twitter and adopts RGCN [27] to learn user representations. RGT [12] introduces relational graph transformers to learn varying influences between different edge relations in heterogeneous graphs. BotMoe [14], the recent one, employs a community-aware mixture-of-experts architecture to capture various patterns in communities in bot detection.

*B. Graph Neural Networks*

**heterophilic Graph Learning.** GNNs [30], [35], [37] are widely studied to aggregate neighborhood features along with graph structure in graph representation learning [47], [48], [49]. Due to this neighborhood aggregation mechanism, many studies [25], [32], [38] posit that most GNNs implicitly assume strong homophily, making them less suitable for capturing heterophilic patterns. Several works [21], [50] even find that GNNs are inferior to MLPs in certain scenarios. Some works [20], [21] extend homophily to "similar neighborhood patterns" assumption. Recent works have developed GNN models aware of heterophilic graphs, such as Geom-GCN [51], H2GCN [32], and GPR-GNN [38]. In this paper, BSG4Bot considers homophily ratios in the subgraphs construction, which boosts both the performance and scalability of bot detection.

**Graph sampling methods.** Graph sampling is a common strategy for training a GNN model over large graphs. Methods like GraphSAGE [37] employ neighborhood sampling, where a random subset of a node's neighbors is selected for aggregation, thereby reducing the computational complexity. Cluster-GCN [18] and AdClusterGCN [19] employ graph clustering algorithms, such as METIS [52], to divide the entire graph into multiple clusters, and then perform training on sampled subgraphs by randomly assembling clusters. GraphSAINT [53] introduces a sampling-based inductive learning framework for generating subgraphs via random walks, node sampling, or edge sampling. NRW [54] presents a method for sampling node pairs in large graphs, aiming to achieve a balance between sampling efficiency and accuracy in large-scale graph analysis. Different from existing graph sampling methods, BSG4Bot considers the homophily ratio in the construction of subgraphs using a pre-trained coarse classifier.

## VI. CONCLUSION

Bot detection has become a hot research topic recently, and the adversarial contest between bot manipulation and detection will continue. Graph-based bot detection is promising as GNN models can leverage content, temporal and topological features naturally. In this paper, we further extend the graph-based bot detection methods, and propose a method named BSG4Bot to build a biased subgraph with enhanced homophily, allowing the GNN model can be trained effectively and efficiently. The long-term features to distinguish the bots and genuine users are explored and incorporated into BSG4Bot. The final experimental results illustrate that BSG4Bot achieves better performance than the state-of-the-art methods, while requiring much less training time.